\documentclass{bmvc2k}

\usepackage{amsmath,amssymb,times}

\usepackage{textcomp}
\usepackage{multirow}
\usepackage{algorithm}
\usepackage[noend]{algpseudocode}
\usepackage{rotating}

\usepackage{graphicx}

\DeclareMathOperator*{\argmin}{arg\,min}

\title{A low-power end-to-end hybrid neuromorphic framework for surveillance applications}

\addauthor{Andres Ussa}{lsiacuc@nus.edu.sg}{1}
\addauthor{Luca Della Vedova}{lucadellavr@gmail.com}{1}
\addauthor{Vandana Reddy Padala}{e170171@e.ntu.edu.sg}{2}
\addauthor{Deepak Singla}{singlade@ntu.edu.sg}{2}
\addauthor{Jyotibdha Acharya}{jyotibdh001@e.ntu.edu.sg}{2} 
\addauthor{Zhang Lei Charles}{charleszhang@ntu.edu.sg}{2}
\addauthor{Garrick Michael Orchard}{garrickorchard@gmail.com}{1}
\addauthor{Arindam Basu}{arindam.basu@ntu.edu.sg}{2}
\addauthor{Bharath Ramesh}{bharath.ramesh03@u.nus.edu}{1}

\addinstitution{
 The N.1 Institute for Health\\
 National University of Singapore\\
 Singapore
}
\addinstitution{
 School of Electrical and Electronic Engineering\\
 Nanyang Technological University,\\
 Singapore
}

\runninghead{A. Ussa \emph{et al}\bmvaOneDot}{Low-Power Neuromorphic Vision Framework}

\begin{document}

\maketitle

\begin{abstract}
With the success of deep learning, object recognition systems that can be deployed for real-world applications are becoming commonplace. However, inference that needs to largely take place on the `edge' (not processed on servers), is a highly computational and memory intensive workload, making it intractable for low-power mobile nodes and remote security applications. To address this challenge, this paper proposes a low-power (5W) end-to-end neuromorphic framework for object tracking and classification using event-based cameras that possess desirable properties such as low power consumption (5-14 mW) and high dynamic range (120 dB). Nonetheless, unlike traditional approaches of using event-by-event processing, this work uses a mixed frame and event approach to get energy savings with high performance. Using a frame-based region proposal method based on the density of foreground events, a hardware-friendly object tracking is implemented using the apparent object velocity while tackling occlusion scenarios. For low-power classification of the tracked objects, the event camera is interfaced to IBM TrueNorth, which is time-multiplexed to tackle up to eight instances for a traffic monitoring application. The frame-based object track input is converted back to spikes for Truenorth classification via the energy efficient deep network (EEDN) pipeline. Using originally collected datasets, we train the TrueNorth model on the hardware track outputs, instead of using ground truth object locations as commonly done, and demonstrate the efficacy of our system to handle practical surveillance scenarios. Finally, we compare the proposed methodologies to state-of-the-art event-based systems for object tracking and classification, and demonstrate the use case of our neuromorphic approach for low-power applications without sacrificing on performance.
\end{abstract}

\section{Introduction}
\label{sec:intro}
Real-time object tracking consists of initializing candidate regions for objects in the scene, assigning them unique identifiers and following their transition. It is a common requirement to further perform classification over the tracked object. These capabilities of object tracking and classification are valuable in applications like human-computer interaction \cite{Jacob2003}, traffic control \cite{Hsieh2006}, medical imaging \cite{Meijering2006} or video security and surveillance \cite{Hampapur2005}. Current methodologies for surveillance tasks use standard cameras that acquire images or frames at a fixed rate regardless of scene dynamics. Consequently, background subtraction to retrieve candidate regions-of-interest for tracking is a computationally intensive step, which is also affected by changes in lighting \cite{Piccardi2004}. On the other hand, deployment of cameras with higher frame rate involves a drastic increase in power requirements \cite{Barnich2011}, besides increased demands in memory and bandwidth transmission. Therefore, the frame-based paradigm tends to be intractable for low-power/remote surveillance applications \cite{Basu2018, Cohen2018}.
\par
As an emerging alternative to standard cameras, event cameras acquire information of a scene in an asynchronous and pixel independent manner, where each of them react and transmit data only when intensity changes happen. This provides a stream of events with a very high temporal resolution (microsecond) at low-power (5-14mW), reducing redundancy in the data with improved dynamic range due to the local processing paradigm. In particular, there is no significant need for background modeling, since a static event camera will only generate events corresponding to moving objects, thereby naturally facilitating tracker initialization. All these features are well suited for visual tracking applications but demand the use of algorithms designed to handle asynchronous events.
\par
The dominant approach in the literature for object tracking and detection using neuromorphic vision sensors has been an event-by-event approach \cite{Lagorce2014, Glover2017, Ramesh2019}. The aim of these methods is to create an object representation based on a set of incoming events and updating it dynamically when events are triggered. Although these methods can be effective for specific applications, they often require high parametrization \cite{Lagorce2014, Glover2017} or are not applicable for tracking multiple objects \cite{Ramesh2019}.
\par
Similar to the above works, \cite{Dardelet2018} is an event-by-event approach for object tracking applications that performs a continuous event-based estimation of velocity using a Bayesian descriptor. Another example is \cite{Ramesh2018}, which proposes event-based tracking and detection for general scenes using a discriminative classification system and a sliding window approach. While these methods work intuitively for objects with different shapes and sizes, and even obtain good tracking results, they have not been implemented under a low-power operation requirement.
\par
\begin{figure}
\centering
\begin{tabular}{c}
\bmvaHangBox{{\includegraphics[width=12.5cm]{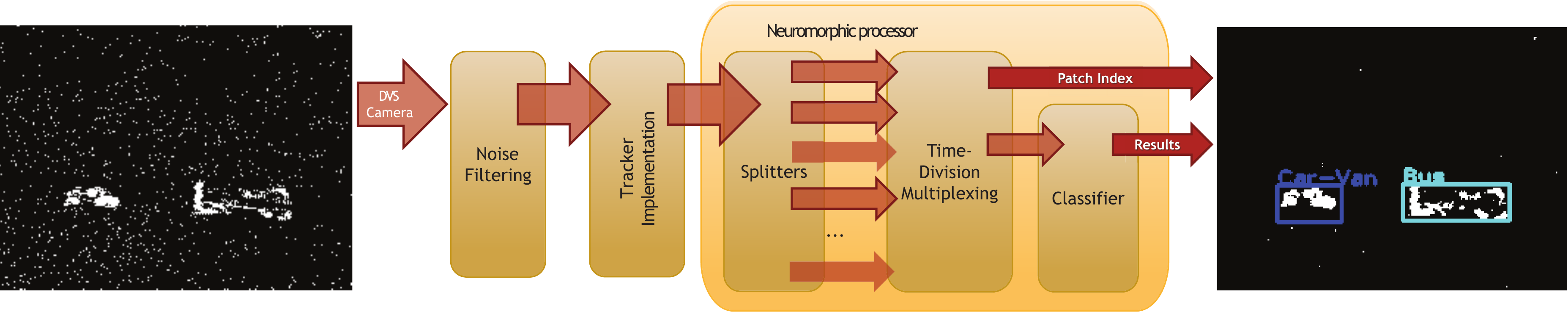}}}
\end{tabular}
\caption{Block diagram of the low-power neuromorphic surveillance system.}
\label{fig:teaser}
\end{figure}

In contrast to the above methods, an aggregation of incoming events can be considered at fixed intervals instead of processing events as they arrive. This produces a more obvious representation of the scene (a ``frame''), and allows an easier coupling with traditional feature extraction and classification approaches \cite{Ni2015, Hinz2017, Iacono2018}. In \cite{Hinz2017}, asynchronous event data is captured at different time intervals, such as 10ms and 20ms, to obtain relevant motion and salient information. Then, clustering algorithms and Kalman filter are applied for detection and tracking, achieving good performance under limited settings. Other examples of event-based frames along with variations in sampling frequency and recognition techniques are \cite{Ni2015, Iacono2018}, which show the potential of this approach for detection. Taking an important step forward for low-power applications, we leverage the low-latency and high dynamic range of event cameras interfaced to an FPGA processor for object tracking, followed by object classification on a neuromorphic chip, to provide an end-to-end neuromorphic framework, as shown in Fig.~\ref{fig:teaser}.
\par
In essence, the focus of our approach is to build a low-power system that takes advantage of a stationary event camera, thereby picking up only moving objects and not being specific to background conditions. To this end, we use a hybrid approach that is different from purely event-based or purely frame-based approaches. First, the asynchronous events are accumulated into a binary image and an overlap-based tracking is performed on these frames. For subsequent object classification, the frames are converted back to spikes for efficient processing on the IBM neuromorphic chip. As shown in the experiments, the hardware-friendly tracker performs significantly better and requires far less resources (7x less memory and 3x less computations) than the popular multi-object event-based mean shift (EBMS) tracker \cite{Delbruck2013}. This is of immense importance when using our fully embedded system for remote surveillance applications where long battery life of the sensor node is critical without sacrificing performance. Next, we describe the details of our framework in the following section, followed by experimental results and conclusions. 

\section{Methodology}
The DAVIS camera events \cite{Brandli2014} are utilized through the formation of frames for the task of tracking vehicles and humans on an urban landscape. Thus, the tracker performance hinges on the ability to capture frames at a rate much faster than the dynamics of the scene, thereby taking advantage of the low-latency of event cameras. The frames obtained are median filtered and region proposals are extracted from two 1-D histograms along $X$ and $Y$ directions for tracking. The tracker uses centroids and Euclidean distances to monitor up to eight objects simultaneously, while classification is performed on these locations using IBM's TrueNorth neuromorphic chip \cite{Merolla2014} to assign one of the following classes: cars, motorbikes, buses, trucks and humans. The filtering and region proposal of the tracker follows the existing method in \cite{Authors} and this work additionally considers occlusion, track velocity calculation and smooth interpolation between two instances of tracking. The full system is embedded on FPGA hardware and interfaced to IBM's TrueNorth chip.
\subsection{Object Tracking}
This work proposes a simple, hardware-friendly tracker operation consisting of a series of steps, namely: track assignment, merging and post-processing. The core function resides in the track assignment task, similar to a Kalman filter update, while the merging and post-processing steps deal with the occlusion and track assignment issues. Each track output is defined by a set of properties: (1) The top-left location of the tracked object (\emph{x} and \emph{y} coordinates); (2) The width and height of the tracked object (\emph{w} and \emph{h}); (3) The velocities $v_x$ and $v_y$ of the object; (4) The tracker state (\emph{free}, \emph{tracking}, or \emph{locked}) and (5) A unique ID. The \emph{free} state indicates that the tracker is in stand-by and no region is currently assigned to it. The \emph{tracking} state indicates that the tracker has matched with a region proposal once. The \emph{locked} state indicates that the tracker has matched a region proposal in at least two consecutive frames and is currently ``locked'' on an object. Since only \emph{locked} trackers are classified, having a \emph{tracking} state allows to filter noisy tracks and reduce the burden on the classifier.
\par
The track assignment procedure can be briefly summarized as follows. As a region proposal, defined by its coordinates, $r_j^{new} = \{x_j^{new}, y_j^{new}, w_j^{new}, h_j^{new}\}$, is received as input, its overlap area with respect to all active trackers $T_j^k = \{x_j^k, y_j^k, w_j^k, h_j^k\}$ will be measured, where $k=1,\cdots,N$ indicates the track IDs and j the frame instance. If their overlap is higher than the \emph{track assignment ratio} $O_{th}$, the region proposal is then assigned to the respective tracker ID. Otherwise, the region proposal is assigned to a \emph{free} tracker. 
\par
To begin with the track assignment, the tracker's new position is estimated based on its previous velocity, and the resulting region is evaluated against the region proposal. The assignment is evaluated based on the calculation of the \emph{track assignment ratio} between two regions, as defined in \eqref{eq:queue1}. 
\begin{multline}
O_{th} = (\max(0, \min(x_j^k + w_j^k, x_j^{new} + w_j^{new}) - \max(x_j^k, x_j^{new})))
\\  \times (\max(0, \min(y_j^k + h_j^k, y_j^{new} + h_j^{new}) - \max(y_j^k, y_j^{new})))
\label{eq:queue1}
\end{multline}
In our implementation, a tracker assignment is made when the overlapping area is higher than 20\%. Subsequently, the tracker properties and state are updated. If the current state is \emph{tracking}, then it is updated to \emph{locked}, and if it was already in \emph{locked} state, it will remain as it is. After a successful assignment, each tracker region is updated using a weighted average as stated in \eqref{eq:queue2}, for each of the spatial elements, where $\alpha$ is the weighting degree coefficient. 
\begin{equation}
T_j^k = (1 - \alpha) \cdot r_j^{new} + \alpha \cdot (T_{j-1}^{k} + v_{j-1}^{k} \cdot \Delta t)
\label{eq:queue2}
\end{equation}
where $v_{j-1}^{k}$ refers to the velocity of the track (in pixels/s) at previous frame instance $j-1$, and $\Delta t = t_{j}-t_{j-1}$. The velocity is calculated as shown in \eqref{eq:vel1}. 
\begin{equation}
\label{eq:vel1}
v_{j}^k(x) = (1 - \alpha) \cdot \frac{(x_j^{new} - x_{j-1}^k) + (w_j^{new} - w_{j-1}^{k})}{\Delta t} + \alpha \cdot v_{j-1}^{k}
\end{equation}
which is also then averaged analogous to the operation performed in \eqref{eq:queue2}. Similarly, velocity calculation is also applicable for the \emph{y} direction.
\par
After the track assignment step, in cases where different region proposals are assigned to the same tracker or vice versa, a merging between the pertinent regions is applied. If more than one region proposal, $r_j^{new_1}$ and $r_j^{new_2}$, is assigned to the same tracker $T_{j-1}^k$, non-maximal suppression is performed among the common regions and the tracker region, to group the rectangles and assign it to $T_{j}^k$. On the other hand, if there is a region proposal that is assigned to more than one tracker, an occlusion check is performed among all the valid trackers.
\par
Considering that the objects to be tracked display a wide range of sizes and often follow opposite directions or move at different speeds, the case in which an object occludes another occurs regularly. In occluded scenarios, the event frame would show a bigger region than the individual objects without a clear boundary between them. In other words, the trackers under evaluation will overlap after one or two steps in the future based on the estimated velocity. Before an occlusion occurs, let us denote the implicated trackers as $T_j^{a}$ and $T_j^{b}$, and their size before occlusion as $(w_o^{a},h_o^{a})$ and $(w_o^{b},h_o^{b})$, respectively. Based on the trackers' original sizes, their velocity, direction and the combined area after occlusion, it is possible to approximate their positions during the occluded frames. For this, a set of conditions is determined: trackers' common direction $cd = v_j^{a} + v_j^b > v_j^a \vee v_j^b$, width increase $wi = (w_j^a > w_{j-1}^{a})$ and highest velocity object $hvo = abs(v_j^{a}) > abs(v_j^{b})$.
\par
While the occlusion is occurring, the change in width of the merged tracks, i.e. $wi$, is used as criteria to determine whether the affected tracks are coming together ($wi=False$) or getting apart ($wi=True$). In particular, for a tracker $T_j^{a}$, the track remains as the current region proposal when $wi$ is $False$, $(x_j^{a}, y_j^{a}, w_j^{a}, h_j^{a})\leftarrow (x_{j}^{new}, y_{j}^{new}, w_{j}^{new}, h_{j}^{new})$, or otherwise when $wi$ is $True$, the track is equal to its original size in the region proposal, $(x_j^{a}, y_j^{a}, w_j^{a}, h_j^{a})\leftarrow (x_{j}^{new}+w_{j}^{new}-w_o^{a}, y_{j}^{new}+h_{j}^{new}-h_o^{a}, w_o^{a}, h_o^{a})$. This situation occurs when the objects are moving in opposite directions ($cd=False$), or when they move in common direction and the velocity of the tracker under evaluation, $T_j^{a}$, is faster than its pair ($hvo=True$). Instead, if $hvo$ is $False$, then the track is intuitively set as the other component of the region proposal, $(x_j^{a}, y_j^{a}, w_j^{a}, h_j^{a})\leftarrow (x_{j}^{new}, y_{j}^{new}, w_o^{a}, h_o^{a})$.
\par
Finally, the post-processing step removes trackers that no longer match any of the region proposals. This is carried out by comparing the \emph{current} state of the tracker with its \emph{past} state. If a tracker was previously set as \emph{locked} or \emph{tracking} state, and in the current frame it no longer exists, then it is likely that the object is lost. However, since a region proposal can be inconsistent through time, due to the hardware noise from the DAVIS, it is important to not free a tracker instantly. An intermediate \emph{maximum unlocks} state is used to determine when a tracker is lost for several consecutive frames, and only then it will be set \emph{free}. Additionally, an out-of-bounds check is performed to release trackers when objects leave the scene.
\par
Note that although the tracking is discontinuous in time, the location and size of the tracked object can be estimated continuously, allowing the size and location of the object to be determined in-between the frames. In other words, any time $t$ satisfying $t_{1}^{k}<t<t_{n}^{k}$,
\begin{equation}
\begin{array}{l l l}
j &= &\underset{t_{i}^{k}>t}{\argmin_i}(t_{i}^{k})\\
\\
\lambda &= &({t-t_{j-1}^{k}})/({t_{j}^{k}-t_{j-1}^{k}})\\
T &= & T_{j-1}^{k} + \lambda(T_{j}^{k}-T_{j-1}^{k})
\end{array}
\end{equation}
where $j$ is the index of the \textit{closest track} and $\lambda$ is the interpolation factor for time $t$. Using the above equations, $T = \{x, y, w, h\}$ will be the location and size of the interpolated \textit{track} at time $t$. This feature will be useful for continuous-time implementations required for certain applications. 

\subsection{Object Classification}
This section describes the process of object classification on the TrueNorth chip using IBM's Energy Efficient Deep Network (EEDN) pipeline. For classifying multiple object tracks, the EEDN pipeline is time-multiplexed to handle eight different objects pseudo-simultaneously. This approach is useful when there are only a few objects to classify in a single frame, as resources (neurons) used for spike generation will scale linearly with the number of objects. However, in general, the input to the TrueNorth chip can either be raw spikes data or images that are subsequently converted to spikes on the host FPGA.
\par
In the case of feeding raw spikes, when TrueNorth is located at a gateway node performing classification on tracks received from multiple sensors, the number of different tracks to classify might be much larger, and therefore, feeding raw spikes would be less feasible to implement. As opposed to continuous streaming of all events, it would be more efficient in terms of time and power to periodically transmit a binary image to TrueNorth. For instance, an uncompressed 32$\times$32 pixel binary image would require a memory of 32$\times$32 bits. In contrast, a continuous spike stream would require a memory of approximately 24 bits per spike. Thus, assuming that each image contains more than $32\times32/24=43$ spikes, the binary image transmission method is more efficient. Note that each TrueNorth classification image is only the tracked regions of the full binary image, which leads to the following integration issues to be considered.
\par
The binary images from each object track are of different sizes, but TrueNorth EEDN requires fixed size images for the CNN classification. Thus, we resize each binary track image to a fixed size, say $42\times42$, before streaming it to TrueNorth. Additionally, EEDN expects a typical RGB image as input, where the first layer of convolution is performed on the host FPGA using multi-bit inputs and the resulting image is thresholded to generate spikes to be fed to the TrueNorth chip. Since we are using 1-bit binary images, we can simply treat any 1s in the image as spikes and perform all convolution layers on TrueNorth, bypassing the need to generate spikes using the host FPGA. This approach is in line with the intended use of IBM's EEDN tools and allows us to better leverage their pipeline. The CNN model we run on TrueNorth is 15 layers deep, similar to the network used in \cite{Esser2015}. 
\begin{figure}[t]
\centering
\begin{tabular}{c}
\bmvaHangBox{\fbox{\includegraphics[width=12cm]{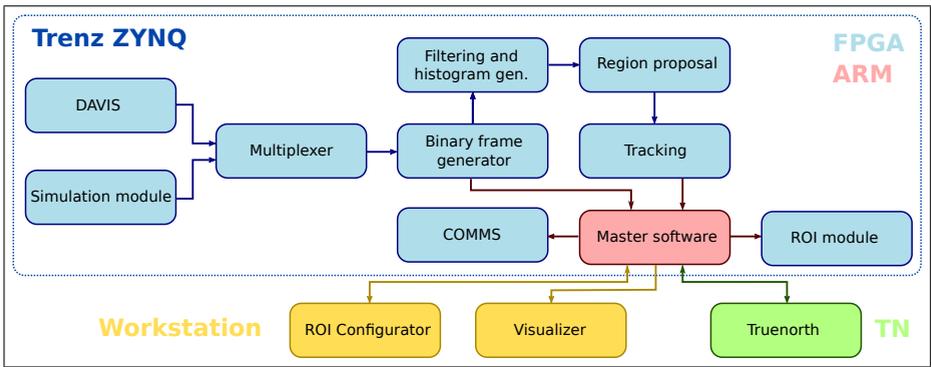}}}
\end{tabular}
\caption{System flow diagram of the end-to-end neuromorphic surveillance system consisting of the DAVIS vision sensor, an FPGA and ARM processor on the Trenz carrier board that is directly interfaced to IBM TrueNorth.}
\label{fig:flowdiag}
\end{figure}
\subsection{Hardware Implementation}
The hardware implementation contains a DAVIS240C event-based sensor, a Trenz TE0720 processing platform that includes an FPGA and an ARM processor, IBM's TrueNorth neuromorphic chip with 4096 cores, and a workstation solely for visualization. The overall operation of the system consists of acquisition of events from the camera, the processing of these events to extract tracked objects, the transmission of these tracked regions to the neuromorphic chip for classification, and the object detection visualization, as shown in Fig.~\ref{fig:flowdiag}.  
\par
The Trenz TE0720 module includes an ARM dual-core Cortex-A9 processing system that runs Ubuntu Linux 16.04 LTS and it handles the interface between the FPGA, the TrueNorth and the workstation for visualization. The communication (COMMS) module implemented on the FPGA allows retrieving the tracker information and sending it to the visualizer after streaming it to TrueNorth for obtaining the classification result. The simulation interface allows sending events from binary files to verify the behavior of the developed modules with prerecorded data. 
\subsubsection{Power Consumption}
The DAVIS can operate at a few milliwatts (10mW), the Trenz operates at about 2W including the base power for running Ubuntu, and the TrueNorth chip operates at 100mW plus 3W for the NS1e evaluation platform. Overall, the power consumption of our system is about 5.2W, which is 40x lower than the latest Nvidia Titan RTX used for conventional deep learning. On the other hand, Brix embedded systems (Intel-I7 processor with 8G RAM), like the one described in \cite{Smedt2015} for real-time object tracking using frame-based sensors, consume about 22W, which is 4x more than our implementation. Note that the Trenz Zynq module is a powerful and flexible development tool, but far exceeds the utilities compared to SmartFusion FPGAs that allow sleep modes, non-volatile configuration memory, and have much lower power consumption overall. In other words, there is significant room for very low-power implementation of our framework with appropriate hardware choices and development efforts. 
\section{Experiments}
The development of this work demanded the acquisition of event-based data from a real application scenario for the purposes of training, validation and testing of the system. The main requirement for these recordings was a high, perpendicular view from the road near intersections. Under this condition, three places inside our campus were chosen for data recording (sample recordings shown in the demo video later). The events are aggregated to generate a frame either every 33 ms or 66ms, and surprisingly even longer time periods (100ms) did not degrade system performance.
\par
Table~\ref{tab:data_overview} shows the distribution of the collected dataset in terms of the number of object tracks for each category. We see that there are lots of car tracks, and thus to balance out the training data, we augmented the dataset with new training samples, created by random flipping, rescaling (up to 140\%) and rotating (up to 20 degrees in either direction). After augmenting a sample, it is cropped back to 42x42 pixels for training with a fixed image size on TrueNorth. A separate test dataset captured at a different time is used for evaluating the system. Manual annotation was carried out to facilitate tracker and classifier evaluation.
\begin{table}[t]
\caption{Number of object tracks per category in the collected dataset.}
\centering
\begin{tabular}{rrrrrrr}
\hline\hline
~ &      Car & Bus   & Pedestrian & Bike & Truck/Van & Total \\[0.5ex]
\hline
Site 1 & 322     & 30    & 115        & 43   & 18    & 528   \\
Site 2 & 226     & 105   & 53         & 14   & 28    & 426   \\
Site 3 & 390     & 181   & 89         & 39   & 56    & 755   \\
Sum & 938     & 316   & 257        & 96   & 102   & 1709  \\
\%  & 54.89   & 18.49 & 15.04      & 5.62 & 5.97  &       \\[1ex]
\hline
\end{tabular}
\label{tab:data_overview}
\end{table}
\begin{table}[t]
\caption{TrueNorth (TN) classification and state-of-the-art DART framework performance.}
\centering
\begin{tabular}{lrrrrrr}
\hline\hline
Method &      Car\% & Bus\%   & Human\% & Bike\% & Truck\% & Overall\%\\[0.5ex]
\hline
TN(Per-Sample)  & 90.4 & 92.5  & 94.7  & 86.9 & 54.2  & 83.3       \\
TN(Per-Track)  & 99.0   & 98.2  & 100   & 100  & 53.8  & 90.2       \\
DART (Per-Track)  \cite{Ramesh2019} & 96.3     & 96.6   & 100         & 100   & 83.9    & 95.4   \\[1ex]
\hline
\end{tabular}
\label{tab:test_accuracy_GT}
\end{table}
\begin{table}[t]
\caption{Type of Training Data vs. Per-Sample TrueNorth Test Accuracy.}
\centering
\begin{tabular}{lrrrrrrr}
\hline\hline
Train Data &      Car      & Bus   & Human & Bike & Truck & Other & Overall\\[0.5ex]
\hline
Ground Truth (\%) & 62.0     & 53.1  & 91.8  & 62.2 & 22.7  & 0   & 48.7     \\
Augmented GT (\%)  & 56.7    & 54.0  & 96.3  & 72.9 & 35.1  & 0   & 52.5    \\
Augmented FPGA (\%)  & 89.8  & 84.2  & 85.8  & 50.0 & 28.5  & 83.9   & 70.4   \\[1ex]
\hline
\end{tabular}
\label{tab:test_accuracy}
\end{table}
\subsection{Results}
Table~\ref{tab:test_accuracy_GT} shows the TrueNorth test accuracy evaluated using the ground truth (not using the tracker output) under two settings:  per-sample (each instance of an object) and per-track (majority voting across all instances of the same object). As expected, there is a significant improvement in the accuracy when considered on a per-track basis. Nonetheless, the TrueNorth model struggles with Trucks due to their similarity to both buses and cars, but does very well on all other classes, especially when given multiple opportunities to classify them on a per-track basis. 
\par
However, it is imperative that for a real-world surveillance application, the back-end classifier is trained on representative samples from the tracker output rather than on manually annotated ground truth tracks\footnote{The tracker output does not have object labels, and in order to train the TrueNorth classifier, class labels are automatically generated using their overlap with the ground truth object locations. Incidentally, spurious tracks with no ground truth overlaps can be labelled as an additional background class and subsequently TrueNorth can classify them as false positives during deployment.}. Table~\ref{tab:test_accuracy} shows the TrueNorth classification accuracies per-sample obtained on the test track output (not the ground truth test tracks) by three models trained on: (1) ground truth track outputs, (2) augmented ground truth track outputs, and (3) the hardware tracker outputs, using the training data. We see that models trained on ground truth tracks that contain no background class perform poorly. This is an important highlight of our proposed system, as mentioned earlier in the introduction, to respond to moving background conditions or spurious tracker output. In particular, the model trained on the tracker output gets a much higher accuracy (70.4\%) compared to models trained on ground truth tracks.
\par
In practice, the above-mentioned system accuracy will be higher due to limitations in auto-generating class labels for the tracker output using the manually annotated ground truth tracks. Trackers that are slightly off their target can only be labelled as background and this is especially true for the case of an object leaving/entering the scene when manual annotations do not exist. Thus, when the tracker locks on before the ground truth tracks have started, TrueNorth correctly classifies it as a bus, but since it does not agree with the annotation, it is marked as an error in the evaluation (Table \ref{tab:test_accuracy}). Similarly, as the object exits the scene, the tracker stops only after the ground truth, so while TrueNorth is again correctly classifying it as a bus, it mismatches with the annotation and is counted as an error. A demo of our full system reveals this scenario clearly\footnote{Video demo: \url{https://tinyurl.com/yyj9k7n2}}.
 
\begin{figure}
\centering
\begin{tabular}{cc}
\bmvaHangBox{\includegraphics[width=60mm]{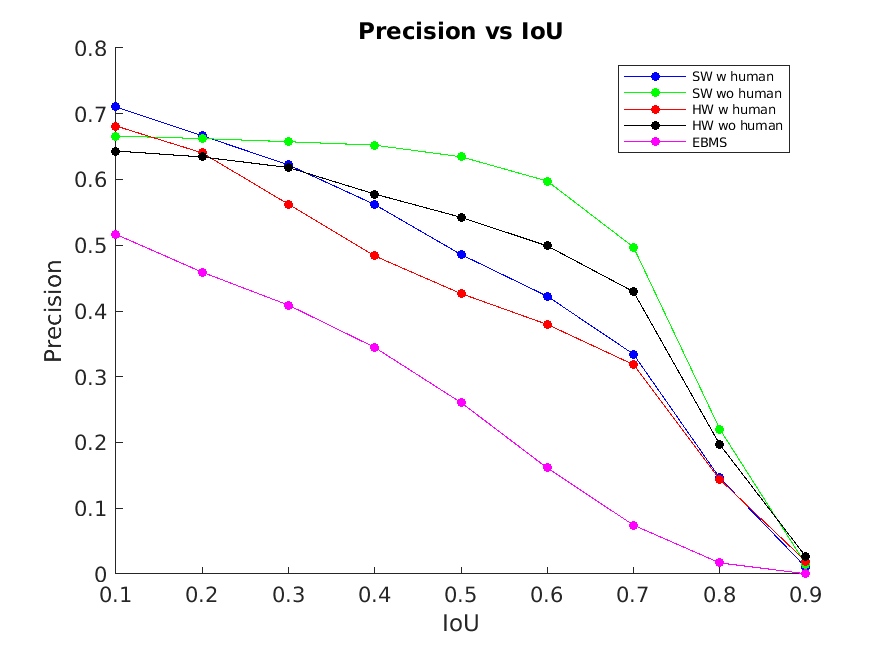}}&
\bmvaHangBox{{\includegraphics[width=60mm]{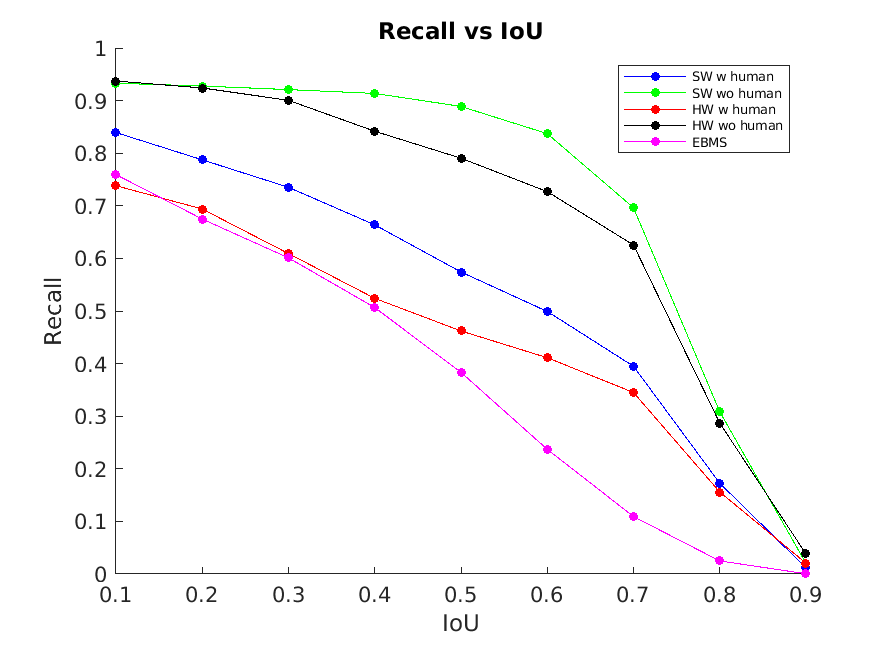}}}\\
(a)&(b)
\end{tabular}
\caption{Precision and Recall plots for the proposed tracker in software (SW), hardware (HW) and EBMS tracker \cite{Delbruck2013} (best viewed on monitor).}
\label{fig4:all_cases_PR}
\end{figure}

\subsection{Comparison to state-of-the-art}
Table~\ref{tab:test_accuracy_GT} compares the TrueNorth classification accuracy to the DART framework \cite{Ramesh2019}, which obtains the state-of-the-art accuracy on multiple event-based object datasets. It is worth stating that the DART method utilizes all the event information and obtains 95.4\% per-track accuracy on the ground truth test dataset. This is understandably higher than the per-track 90.2\% accuracy obtained using the EEDN framework, but more importantly shows that event cameras can be utilized to generate ``frames'' without sacrificing much performance for low-power surveillance applications. 
\par
Fig.~\ref{fig4:all_cases_PR} shows the performance of the tracker in software, hardware and the popular multi-object event-based mean-shift (EBMS) tracker \cite{Delbruck2013}. To analyze the effect of finite bit precision in hardware implementations, we have extracted the region proposals from the FPGA and passed it through the software tracker to generate precision and recall curves as shown in Fig.~\ref{fig4:all_cases_PR}. All the curves are generated by sweeping the intersection over union (IoU), a common metric used to evaluate the quality of match between ground truth and tracker output, following the protocols in \cite{Authors}. It was observed that the tracker performance was much worse for humans than other objects due to their smaller size and velocity. Hence, we show the results for the tracker with and without humans. At an IoU of 0.4, keeping humans in the analysis, the precision and recall values of the hardware tracker are 0.48 and 0.52 respectively. After eliminating the human tracks, the precision and recall values jump to 0.58 and 0.84 respectively. Additionally, it can be seen that the hardware tracker outperforms the  multi-object EBMS tracker \cite{Delbruck2013} comfortably (7x less memory and 3x less computations \cite{Authors}) while following a similar trend as the software version with around 5\% drop in accuracy due to finite precision arithmetic.



\section{Conclusion}
This paper presented one of the first end-to-end neuromorphic frameworks for real-world object tracking and classification demonstrated using a low-power hardware implementation that consumes about 5W, which is 40x less power than conventional GPUs used for deep learning and 4x lesser than state-of-the-art frame-based embedded systems for real-time tracking. The proposed framework employs a hybrid approach consisting of events aggregated into frames for maintaining individual track of objects in occluded scenarios, and subsequently the tracked object was efficiently classified using the IBM EEDN pipeline of the spike-based neuromorphic chip. In this setup, the TrueNorth chip was time-multiplexed to handle eight objects while making sure that the neurons used for pre-processing will scale linearly with the number of objects to be pseudo-simultaneously classified. In addition to a real-time demo, we compared the proposed tracking and classification methods to state-of-the-art event-based methods and showed its relevance to on-going work in the research field. In summary, we have demonstrated a strong use case of our neuromorphic framework for low-power applications without sacrificing on performance.

\section*{Acknowledgements}
The authors wish to thank Zhang Shihao for collecting data and asssiting in annotations that enabled quantitative evaluation. Additionally, the authors would like to thank Cedric Seah for setting up a preliminary tracking system using a C++ real-time implementation. 

\bibliography{sig-alternate-sample}
\end{document}